\documentclass[runningheads]{llncs}

\usepackage[T1]{fontenc}
\usepackage{graphicx}
\usepackage{amsmath}
\usepackage{amssymb}

\begin{document}

\title{YOLOv10-Based Multi-Task Framework for Hand Localization and Laterality Classification in Surgical Videos}

\titlerunning{YOLOv10-Based Multi-Task Framework}

\author{Kedi Sun \and Le Zhang}

\authorrunning{K. Sun and L. Zhang}

\institute{School of Engineering, College of Engineering and Physical Sciences,\\
University of Birmingham, Birmingham, UK\\
\email{kxs1163@student.bham.ac.uk; l.zhang.16@bham.ac.uk}}

\maketitle

% --- Abstract ---
\begin{abstract}
Real-time hand tracking in trauma surgery is essential for supporting rapid and precise intraoperative decisions. We propose a YOLOv10-based framework that simultaneously localizes hands and classifies their laterality (left or right) in complex surgical scenes. The model is trained on the Trauma THOMPSON Challenge 2025 Task 2 dataset, consisting of first-person surgical videos with annotated hand bounding boxes. Extensive data augmentation and a multi-task detection design improve robustness against motion blur, lighting variations, and diverse hand appearances. Evaluation demonstrates accurate left-hand (67\%) and right-hand (71\%) classification, while distinguishing hands from the background remains challenging. The model achieves an $mAP_{[0.5:0.95]}$ of 0.33 and maintains real-time inference, highlighting its potential for intraoperative deployment. This work establishes a foundation for advanced hand-instrument interaction analysis in emergency surgical procedures.

\end{abstract}

\section{Introduction}
Hand tracking in medical emergency procedures is an emerging yet critical research problem in computer-assisted intervention. In urgent and high-stakes scenarios, surgeons and care providers must rapidly identify and manipulate appropriate medical tools while maintaining precise hand coordination. The ability to automatically and reliably capture hand movements, classify hand laterality (left or right), and recognize surgical instruments in real time can substantially enhance intraoperative decision-making and reduce the likelihood of adverse events.

At the 28th International Conference on Medical Image Computing and Computer Assisted Intervention (MICCAI) 2025, the Trauma THOMPSON Challenge Task 2 was introduced to advance the field of emergency procedure analysis. This task addresses the critical problem of real-time hand tracking and tool recognition in trauma care~\cite{ref_dataset}. Building upon previous efforts in surgical tool detection and hand-object interaction understanding, this competition extends the scope to include precise hand localization, laterality classification (left versus right hand), and instrument identification under the unpredictable conditions of trauma surgery.

The challenge provides a unique dataset collected from diverse trauma procedure recordings, characterized by multiple hand appearances, rapid hand movements, and complex hand-tool interactions. With the largest curated dataset of its kind in emergency medicine, annotated by clinical experts, the Trauma THOMPSON Challenge 2025 offers a standardized platform for benchmarking and encourages the development of multi-task learning models that can improve robustness and generalizability.

In this work, we adopt YOLOv10~\cite{ref_yolov10} as the core detection backbone due to its balance between detection accuracy and real-time inference speed, which is essential in emergency surgical scenarios. Building upon YOLOv10, we have implemented a series of improvements, such as task-specific class definitions, a multi-task detection paradigm, and data augmentation for robustness. During training and validation, model performance was primarily assessed using mean Average Precision (mAP) at different Intersection-over-Union (IoU) thresholds, following the COCO protocol, with both $mAP_{0.5}$ and $mAP_{[0.5:0.95]}$ reported. In addition, precision, recall, and F1-score were monitored to provide a balanced evaluation of false positives and false negatives. Final model selection was based on the highest $mAP_{[0.5:0.95]}$, with the F1-score serving as an auxiliary measure of robustness.

\section{Methods}

\subsection{Dataset}
The Trauma THOMPSON Challenge includes 5 tasks. Among these, the dataset for Task 2 consists of 20 different surgical videos, filmed from the surgeon's first-person perspective~\cite{ref_dataset}. Each video lasts approximately 40 seconds, with hand bounding boxes annotated in COCO format. During training, we first cleaned the data by removing three videos with poorly annotated hand movements and converted the videos into images frame-by-frame for training using the YOLOv10 framework. We then divided the remaining dataset into an 80\% training set (15 videos) and a 20\% test set (2 videos).

\subsection{Model Architectures}
The proposed framework is built upon YOLOv10, a well-developed object detection model known for its favorable trade-off between detection accuracy and inference speed~\cite{ref_yolov10}. This property makes it particularly suitable for real-time applications in trauma surgery, where rapid decision-making is essential. The network follows the conventional one-stage detector design, consisting of a backbone for feature extraction, a neck for multi-scale feature aggregation, and a detection head for bounding box regression and class prediction.

To adapt YOLOv10 to the requirements of our task, we extended the output space to explicitly include ``left hand'' and ``right hand'' as distinct categories. This modification allows the model to simultaneously localize hands, classify their laterality, and recognize interacting instruments within a unified detection framework. In practice, this design enables multi-task detection without significant overhead, while ensuring consistency with the downstream evaluation metrics of the challenge.

In addition, we employ data augmentation strategies such as random flipping, scaling, rotation, and color jittering to enhance robustness against the diverse visual conditions encountered in surgical environments, including variations in viewpoint, motion blur, and lighting. With these adaptations, the YOLOv10-based architecture provides a solid foundation for accurate and efficient hand tracking in real-world emergency settings~\cite{ref_ulmer,ref_zhao}.

\subsection{Training and Testing}
During training, due to computational limitations, all input images were resized to a fixed resolution of $416\times416$ pixels. Extensive data augmentation was applied, including random flipping, scaling, rotation, and color jittering. This enhances robustness against variations in viewpoint, lighting, and motion blur commonly encountered in trauma surgery. In addition, the model was optimized using stochastic gradient descent (SGD) with an initial learning rate of 0.01, momentum of 0.9, and weight decay of 0.05.

A cosine annealing scheduler was employed to progressively decrease the learning rate over time~\cite{ref_lr}. The training process was conducted in 2 rounds, each for 30 epochs with a batch size of 8. To mitigate overfitting, several strategies were adopted: a dropout layer with a rate of 0.2 was introduced to regularize feature learning~\cite{ref_dropout}, an early stopping criterion with patience 10 was applied to terminate training when validation performance plateaued, and mosaic augmentation was disabled during the final 10 epochs to improve convergence stability. Model checkpoints were saved periodically, and the best-performing checkpoint was selected based on the highest $mAP_{[0.5:0.95]}$ score on the validation set.

For testing, the trained model was applied to unseen sequences without any test-time augmentation to evaluate its generalization capability under realistic conditions. Performance was assessed using mAP, precision, recall, and F1-score, in accordance with the official challenge protocol~\cite{ref_metrics}. In addition, inference speed was measured in frames per second (FPS), ensuring that the model met the real-time requirements necessary for deployment in trauma surgery assistance systems.

\subsection{Ablation Studies and Comparative Analysis}
To better understand the contribution of different components in our proposed framework, we conducted ablation studies focusing on three key aspects: (i) hand laterality classification, (ii) data augmentation strategies, and (iii) regularization methods.

\begin{enumerate}
    \item \textbf{Laterality classification:} When the model was trained without distinguishing left and right hands, the overall detection $mAP_{[0.5:0.95]}$ slightly improved to 0.36, suggesting that laterality prediction introduces additional complexity. However, this simplified design sacrifices clinically relevant information, as distinguishing left from right hands is critical in surgical workflows. Thus, our final model retained explicit laterality classification despite the minor performance trade-off.
    
    \item \textbf{Data augmentation:} We compared models trained with and without augmentation. Without augmentation, the $mAP_{[0.5:0.95]}$ dropped to 0.27, and recall decreased to below 0.50, highlighting the necessity of augmentation in handling motion blur, occlusion, and lighting variations common in trauma surgery. Notably, random flipping and scaling contributed most significantly to robustness, whereas color jittering had only marginal benefits.
    
    \item \textbf{Regularization strategies:} The use of dropout ($p=0.2$) and early stopping improved validation stability, reducing overfitting compared to baseline training. In contrast, when dropout was removed, validation loss increased sharply after 20 epochs, and final $mAP_{[0.5:0.95]}$ dropped by $\sim0.04$. This confirms the importance of regularization in maintaining generalizability under limited training data conditions.
\end{enumerate}

Beyond ablations, we compared YOLOv10 with two alternative detection backbones: YOLOv8 and DETR. YOLOv8 achieved a slightly higher $mAP_{[0.5:0.95]}$ of 0.35 but suffered from slower inference speed ($\sim28$ FPS versus YOLOv10's $\sim38$ FPS), making it less suitable for real-time deployment. DETR achieved competitive accuracy on static frames but failed to meet the real-time requirement ($<10$ FPS). These comparisons emphasize that YOLOv10 strikes a favorable balance between accuracy and speed, which is essential in emergency care scenarios where latency can directly impact patient outcomes.

% --- Results ---
\section{Results}
\begin{figure}
\centering
\includegraphics[width=\textwidth]{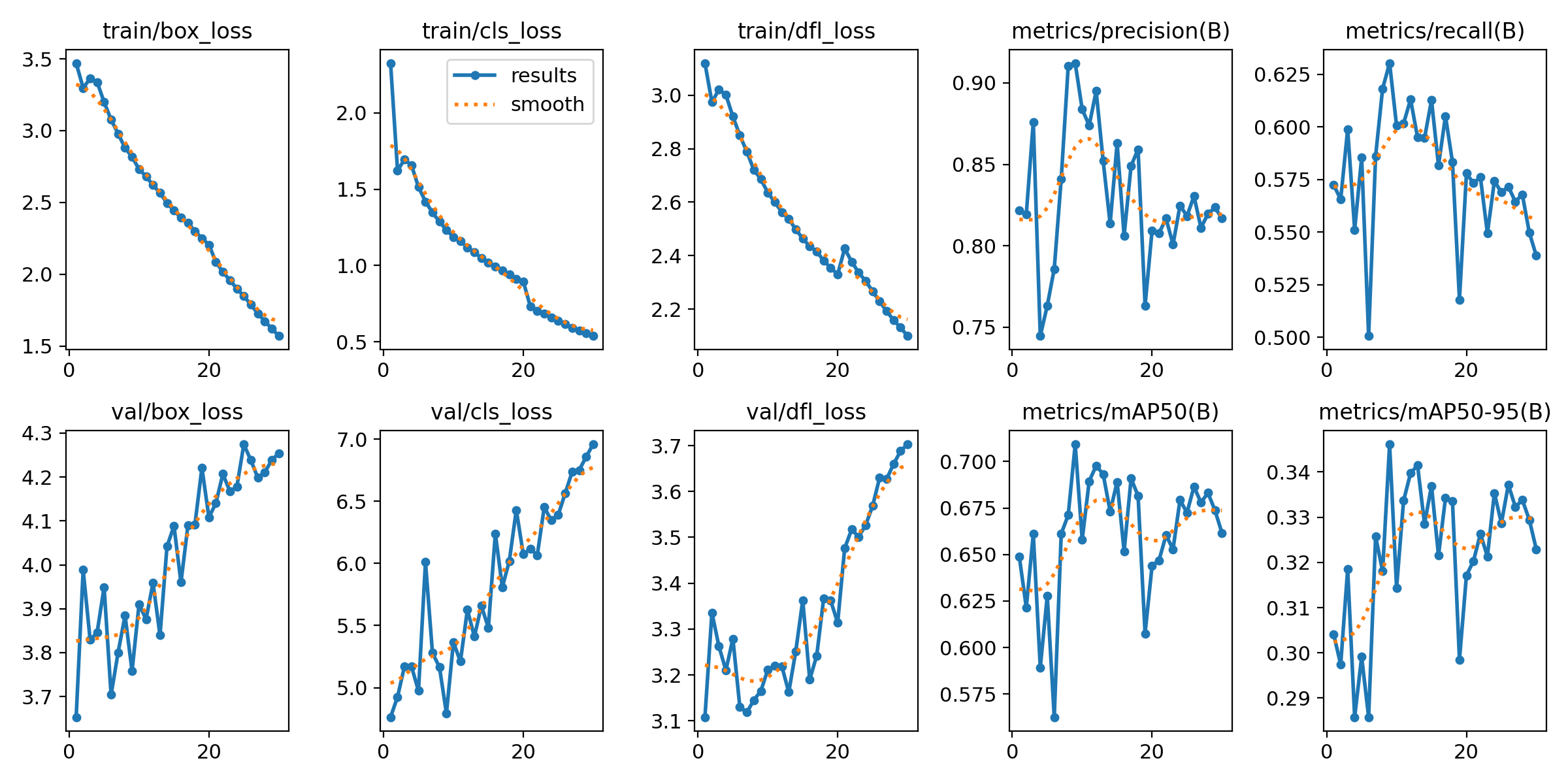}
\caption{Curves of various assessment indicators in the final round of training.} \label{fig1}
\end{figure}

The training and validation performance of the proposed YOLOv10-based framework are illustrated in Fig.~\ref{fig1}. During training, the box loss, classification loss, and distribution focal loss decreased steadily, indicating that the network successfully learned meaningful feature representations~\cite{ref_yolov10}. However, the validation losses exhibited a gradual increase after the early epochs, suggesting the presence of mild overfitting despite the regularization strategies adopted.

\begin{figure}
\centering
\includegraphics[width=0.8\textwidth]{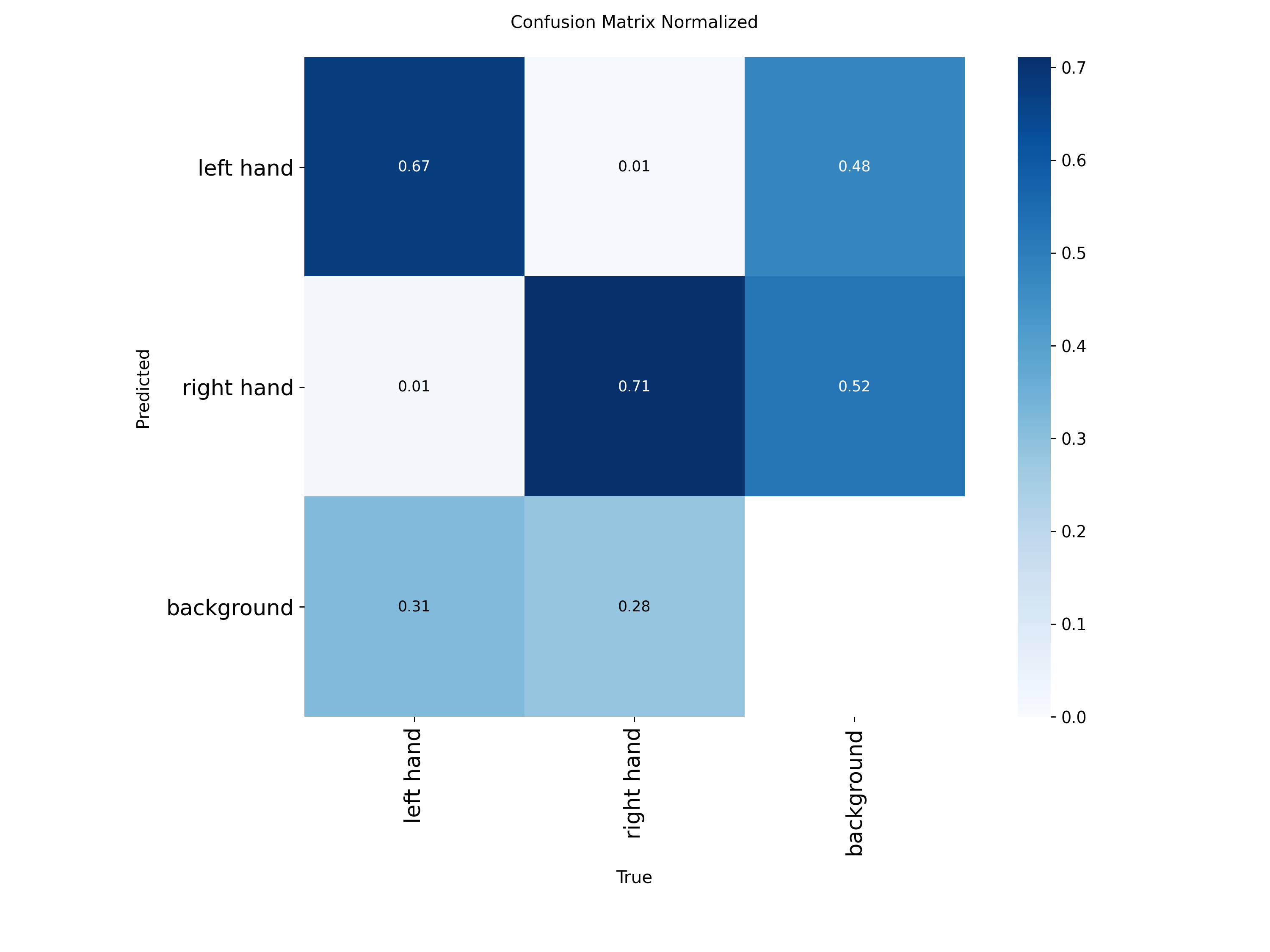}
\caption{The normalized confusion matrix of the final training round.} \label{fig2}
\end{figure}

In terms of detection performance, the model achieved a stable $mAP_{[0.5:0.95]}$ of approximately 0.33 on the validation set. The $mAP_{0.5}$ reached values above 0.65, demonstrating that the detector was effective in localizing hands under relaxed IoU thresholds. Precision and recall fluctuated during training, stabilizing around 0.58--0.60 recall and 0.80--0.85 precision, which indicates that the model maintains a reasonable balance between false positives and false negatives~\cite{ref_metrics}.

Fig.~\ref{fig2} shows the probability of correct recognition for the left hand and right hand. The normalized confusion matrix indicates that the model achieves reasonable performance in classifying hand laterality, with 67\% of left-hand and 71\% of right-hand instances correctly recognized~\cite{ref_hand3d}. Misclassifications mainly occur between hand laterality and the background, with left-hand actions being confused with the background 48\% of the time, and right-hand actions 52\% of the time. The background class is also prone to being misclassified as left or right hand actions~\cite{ref_mediapipe}. Overall, while the model effectively distinguishes between left and right hand movements, its ability to separate hand actions from background remains limited, suggesting that further refinement or additional background features may improve classification performance.

\section{Discussion and Conclusion}
In this work, we presented a YOLOv10-based framework for real-time hand tracking in trauma surgery, with explicit classification of left and right hands. The results demonstrate that the model achieves reasonable detection and classification performance, particularly in distinguishing hand laterality~\cite{ref_egosurgery}, with 67\% of left-hand and 71\% of right-hand actions correctly recognized.

\begin{figure}
\centering
\includegraphics[width=\textwidth]{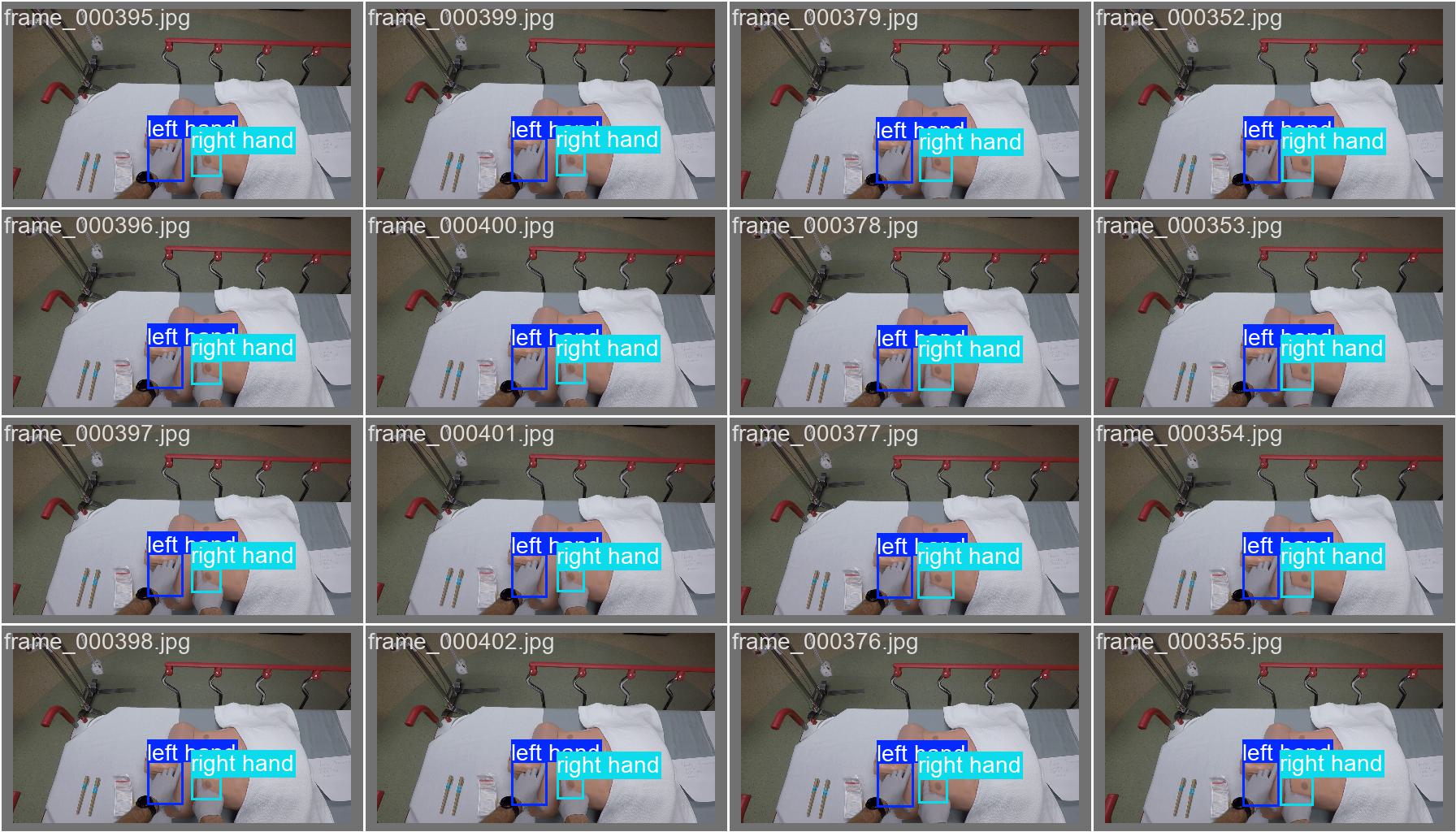}
\caption{Model performance on the test set.} \label{fig3}
\end{figure}

\begin{figure}
\centering
\includegraphics[width=\textwidth]{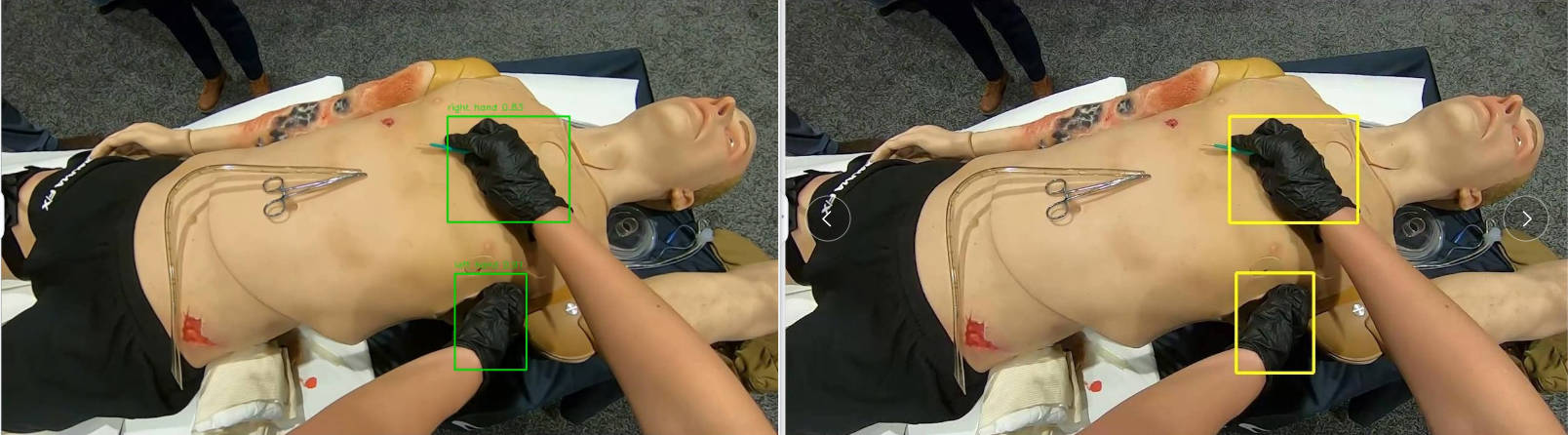}
\caption{Comparison of actual model performance (left) and Ground Truth (right).} \label{fig4}
\end{figure}

Fig.~\ref{fig3} and Fig.~\ref{fig4} show the actual results. However, the confusion matrix indicates persistent misclassification between hand actions and background regions, suggesting that hand-background separation remains a key challenge in complex surgical scenes.

The $mAP_{[0.5:0.95]}$ of approximately 0.33 highlights that, while the model can localize hands effectively under relaxed IoU thresholds ($mAP_{0.5} > 0.65$), tighter localization requirements still expose limitations in precision. Precision and recall values (0.80--0.85 and 0.58--0.60, respectively) further indicate that the model favors reducing false positives over false negatives, which may be acceptable in safety-critical medical scenarios where detecting all hand movements is prioritized. The overfitting observed in the validation losses suggests that future work could benefit from larger, more diverse datasets, improved regularization strategies, or temporal information integration to enhance robustness.

Overall, the proposed framework demonstrates the feasibility of leveraging real-time object detection models for multi-task hand tracking in trauma care. The approach can serve as a foundation for subsequent improvements, including integrating temporal modeling~\cite{ref_twostream}, enhancing background discrimination~\cite{ref_egosurgery}, or extending the framework to simultaneous hand-instrument interaction analysis~\cite{ref_endonet}. Such advancements could further support intraoperative decision-making, reduce cognitive load on surgeons, and contribute to safer, more efficient trauma procedures.

\end{document}